\documentclass[10pt,twocolumn,letterpaper]{article}

\usepackage{iccv}
\usepackage{times}
\usepackage{epsfig}
\usepackage{graphicx}
\usepackage{amsmath}
\usepackage{amssymb}
\usepackage{multirow}
\usepackage{dblfloatfix} 
\usepackage[mathcal]{euscript}

\usepackage[pagebackref=true,breaklinks=true,letterpaper=true,colorlinks,bookmarks=false]{hyperref}

\iccvfinalcopy 

\def\iccvPaperID{19} 

\ificcvfinal\fi

\begin{document}

\title{Craquelure as a Graph: Application of Image Processing and Graph Neural Networks to the Description of Fracture Patterns}

\author{Oleksii Sidorov and Jon Hardeberg\\
The Norwegian Colour and Visual Computing Laboratory, NTNU\\
Gj\o vik, Norway\\
{\tt\small oleksiis@stud.ntnu.no, jon.hardeberg@ntnu.no}
}

\maketitle
\ificcvfinal\thispagestyle{empty}\fi

\begin{abstract}
Cracks on a painting is not a defect but an inimitable signature of an artwork which can be used for origin examination, aging monitoring, damage identification, and even forgery detection. This work presents the development of a new methodology and corresponding toolbox for the extraction and characterization of information from an image of a craquelure pattern.

The proposed approach processes craquelure network as a graph. The graph representation captures the network structure via mutual organization of junctions and fractures. Furthermore, it is invariant to any geometrical distortions. At the same time, our tool extracts the properties of each node and edge individually, which allows to characterize the pattern statistically. 

We illustrate benefits from the graph representation and statistical features individually using novel Graph Neural Network and hand-crafted descriptors correspondingly. However, we also show that the best performance is achieved when both techniques are merged into one framework. We perform experiments on the dataset for paintings’ origin classification and demonstrate that our approach outperforms existing techniques by a large margin.

\end{abstract}

\section{Introduction}

The recent tragedy of Notre Dame de Paris has shaken the society and reminded how important cultural heritage is. While also being a multi-billion market, fine art is a field of intense research in history, chemistry, imaging, and conservation. In the last decades, it has benefited from digital image processing which is a convenient tool for capture and analysis of visual information.

The majority of the painted or varnished artworks contain a fine network of fractures also known as craquelure. A number of recent works in image processing \cite{giakoumis1998digital}\cite{5968544}\cite{mosaic} address craquelure as an adverse effect of aging and aim to detect and inpaint cracks in the digital copy. Our work, on the contrary, consider crack pattern an additional source of information which is so valuable in historical studies.

The early works in this area present attempts to extract useful information from the craquelure image \cite{abas2002content}\cite{10.2307/1506653}. However, they only cover basic statistical features while our model is the first model which describes the topography of the nodes, shape of the edges, and organization of the network via graph theory. 

Graph representation also allows us to use the state-of-the-art deep learning framework Graph Neural Network for learning the pattern’s features directly from its structure, which, to the best of our knowledge, has never been applied to craquelure. Experimental results of our approach are shown to be very promising, and we believe it will initiate a new discussion in the field.

Additionally, our work fills another big gap – the absence of openly available software for extraction of information from binary or RGB image of craquelure. We publish all source codes, and make it possible to easily extract graphs, calculate statistics, visualize the results, analyze links orientation and nodes topography as well as perform next steps of using extracted features for classification. We believe our code will be beneficial for many conservators and art researchers in their work. 

The rest of the paper is organized in the following way: Chapter 2 covers existing approaches to craquelure description, Chapter 3 discusses the theory and design the proposed approach, whereas Chapter 4 describes experimental details and results.

\section{Related Works}
Cracks appear due to the drying of paint or varnish followed by shrinkage of its volume while being stretched on the base with the fixed area. This process leads to the increase of surface stress and finally tearing the material. The work of Bratasz and Sereshk \cite{doi:10.1080/00393630.2018.1504433} provides a comprehensive description of the physical model of craquelure formation due to the increased stress and shear stress. Whereas work of Krzemien \etal \cite{doi:10.1080/00393630.2016.1140428} describes the environmental conditions responsible for craquelure formation.

At one point of the history it was claimed that "purely verbal or mathematical descriptions of crack patterns are almost impossible" \cite{boers}, luckily, modern technology allows us to make the impossible possible. Bucklow \cite{10.2307/1506709} was the first who proposed a systematic approach to formal description of craquelure and performed analysis of cracking in relation to materials and methods employed by the artist. In later work \cite{10.2307/1506653}, Bucklow proposed a semi-automatic approach based on heuristic representation obtained from human experts as well as an algorithmic representation of a digital image as a set of Bezier curves. The obtained representations were used to classify patterns using Linear Discriminant Analysis and Multi-Layer Perceptron. 

The next step in the development was made by Abas \cite{abas2004}\cite{abas2002content} who used image morphology followed by directional filtering to process the image while preserving spatial information. A set of hand-crafted filters in grid representation was used by him to gather statistics of the fracture orientation cell-wise, and use it to identify distinguishable types of networks and cluster them using K-means clustering. In a modern perspective, the usage of hand-crafted filters and splitting the image into the grid cells is a strong limitation of the method, whereas cracks orientation is just a small part of the information contained in a pattern. 

Similarly, Freeman \etal \cite{freeman2013} studied directionality histogram extracted from skeletonized image but also measured properties of the "islands" -- regions surrounded by cracks. Junctions were studied by Taylor \etal \cite{10.1007/978-3-319-20125-2_15} who extracted junction points using crossing number and used them as features for matching similar craquelure patterns to detect forgery. However, this approach does not take into account links between junctions, neither can it be used as a generic descriptor of a pattern. A more comprehensive model was developed by El-Youssef \etal \cite{el2014development}. It extracts arrays of nodes and edges, and describes each element by its chain code and location. We find such representation inconvenient for algorithmic processing, moreover, authors themselves use only 11 statistical measures when demonstrating the method's application. Use of chain codes and the extraction of the histogram of only four directions limit processing as well.  

Our model covers all mentioned approaches and produces the most comprehensive description of a network as individual properties of nodes and edges as well as their mutual organization. Moreover, unlike the above-mentioned works, we make source code openly accessible\footnote{Source code: \url{https://github.com/acecreamu/craquelure-graphs}}, so the community of conservators and art specialists can benefit from it.

There is also a set of works designed exclusively for cracks detection, \textit{i.e.} extraction of the binary mask of cracking from an RGB image. For this purpose, authors utilize morphology \cite{1980421}, clustering \cite{10.1007/978-3-642-32518-2_35}, Bayesian classification \cite{6622710}, deep learning \cite{sizyakin2018deep}, and other modalities like X-ray \cite{5968544} and hyperspectral imaging \cite{7274902}\cite{8706194}. These approaches may be used as a prerequisite for the proposed algorithm.

\section{Methodology}
Due to the wide variety of effective crack detection algorithms and even wider variety of available imaging technologies, our algorithm takes as an input skeletonized binary image $\mathcal{I} \in R^2$ of a crack network $C=(\mathcal{C},\mathcal{N})$ which consists of cracks $\{c_1,\ldots,c_N\}\in \mathcal{C}$ and nodes $\{n_1,\ldots,n_N\}\in\mathcal{N}$. Additionally, we propose an interface which was used in our experiments to perform basic, yet effective craquelure segmentation using black top-hat (bottom-hat) transform (Eq. 1) followed by adaptive thresholding \cite{doi:10.1080/2151237X.2007.10129236} and area cleaning.
\begin{equation}
    \mathcal{I}=ADAPT.THRESH.(I^-) : I^-=I\,\bullet\, s(r)-I
\end{equation}
where $I$ is a grayscale image, $I^-$ -- the result of transformation with enhanced fractures, sign $\bullet$ denotes morphological closing with structural element $s$ of radius $r$.

The obtained skeletonized image was transformed into a non-directed unweighted graph $G=(V,E)$ where $V=\mathcal{N}$ and $E=E(\mathcal{C})$, using the algorithm of Kollmannsberger \cite{kollmannsberger2017small} improved by removing false nodes near image border\footnote{The first issue occurs due to the intersection of a fracture with an image border. The algorithm marks an intersection point as the end of a fracture, even though the fracture continues beyond the image area. To correct it, we simply disregard the detected nodes at an image border.} and merging falsely separated nodes\footnote{Skeletonization procedure may split one node with a large number of edges onto a few closely located nodes. Thus, we join the nodes located at the distance comparable to the fracture width via Euclidean distance.}. 

\begin{figure}[t]
\begin{center}
  \includegraphics[width=\linewidth]{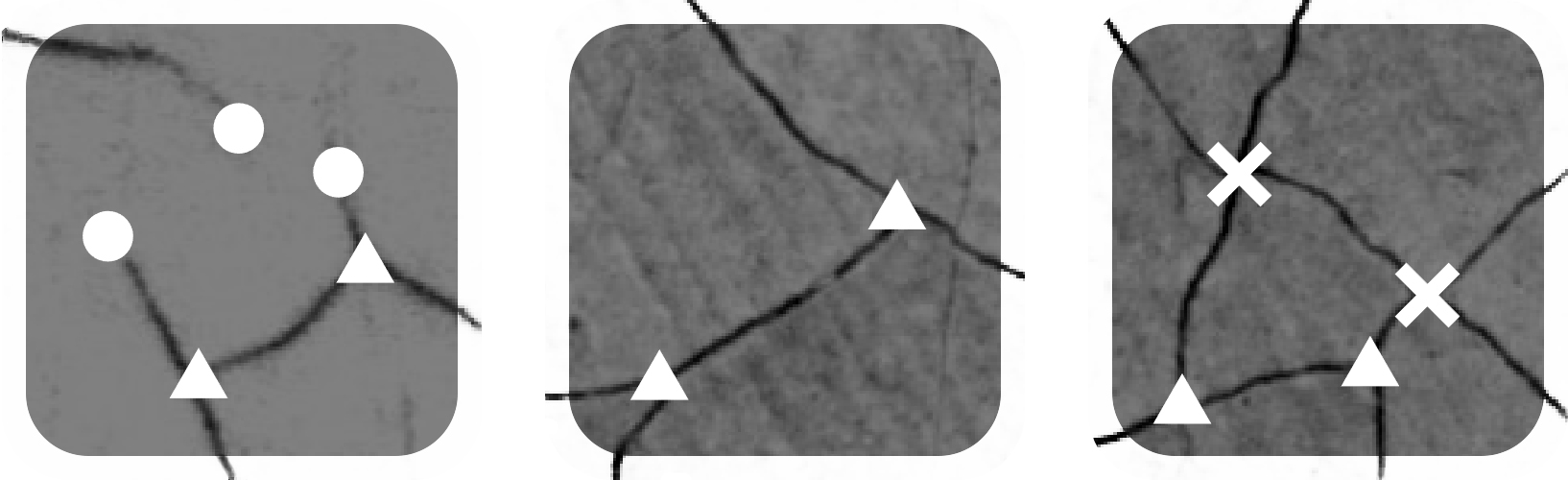}
  \end{center}
  \caption{Illustration of the nodes of O, Y, and X types.}
  \label{fig:oyx}
\end{figure}

\begin{figure*}[t]
\begin{center}
  \includegraphics[width=\linewidth]{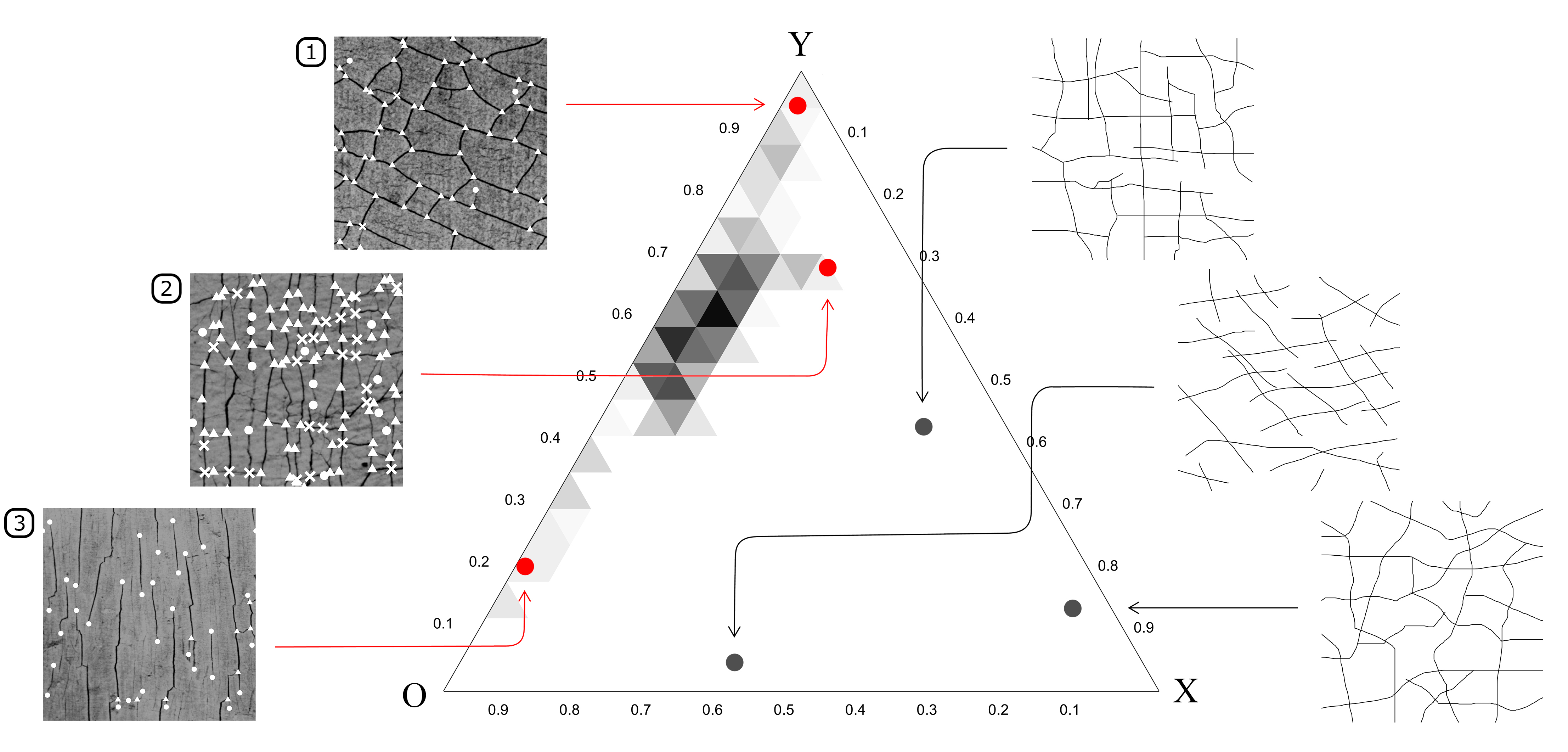}
  \end{center}
  \caption{Triangular topography chart. Each point is defined by three coordinates $(\mathcal{N}_O, \mathcal{N}_Y, \mathcal{N}_X)$. Darkened areas illustrate distribution of natural craquelure network on the chart (darker color –- higher density). Real samples shown: (1) -– XVI century Flemish panel; (2) -– Jan van de Cappelle, "River scene with a large ferry", canvas painting, 1665; (3) –- Duccio, "The Annunciation", panel, 1307/8.}
  \label{fig:triangle}
\end{figure*}

\subsection{Properties of nodes}
Naturally, the intersection points of multiple fractures are equivalent (unlike the atoms in vertices of a molecule, for example) and at most characterized by their amount and location. Here, we propose a novel approach of labeling nodes by their topology, \textit{i.e.}, the number of fractures intersecting at one point, and illustrate its descriptive power. The similar approach is used in the field of geology and tectonics
\cite{sanderson2015use}\cite{SANDERSON2018}.

\begin{table}[b]
\begin{center}
\begin{tabular}{|p{0.5cm}|cccc|}
\hline
& mean & std & 5th percentile & 95th percentile\\
\hline\hline
O	&0.343	&0.135	&0.138	&0.605 \\
Y	&0.592	&0.127	&0.371	&0.829 \\
X	&0.065	&0.035	&0.000	&0.126 \\
\hline
\end{tabular}
\end{center}
\caption{Distribution of the node-types in the real networks.}
\label{tab:node-types}
\end{table}

We denote endpoints of branch fractures which have only one connection as type "O" (marked by a circle), crossing of three cracks as type "Y" (marked by a triangle), whereas junctions of four cracks are labeled as "X" (marked by a square) (Fig. \ref{fig:oyx}). Junctions of higher order are rare in cultural heritage and are included in "X"-type. 

A particular network (graph) can be characterized by the ratio of O, X, and Y nodes $(\mathcal{N}_O, \mathcal{N}_Y, \mathcal{N}_X): \sum_{i=O,Y,X} \mathcal{N}_i =1$. The triangular chart suits such visualization well. Figure \ref{fig:triangle} illustrates how the topology of a network relates to the position on the chart. Darkened areas correspond to the density (2D histogram) of the points extracted from a set of real craquelure patterns on paintings of various origin\footnote{For the data description see Section 4.1.}. It may be seen that the majority of patterns are localized between O and Y axes, whereas X-nodes are uncommon. Table \ref{tab:node-types} presents quantitative description of the distribution and proves rarity of X-nodes in paintings. 

For a better understanding of the chart and topology metric, Fig. \ref{fig:triangle} also contains several examples of real cracks (red points) as well as theoretical simulation of patterns which we did not observe in the experiments (black dots).

\subsection{Properties of edges}
\begin{figure}[t]
\begin{center}
  \includegraphics[width=\linewidth]{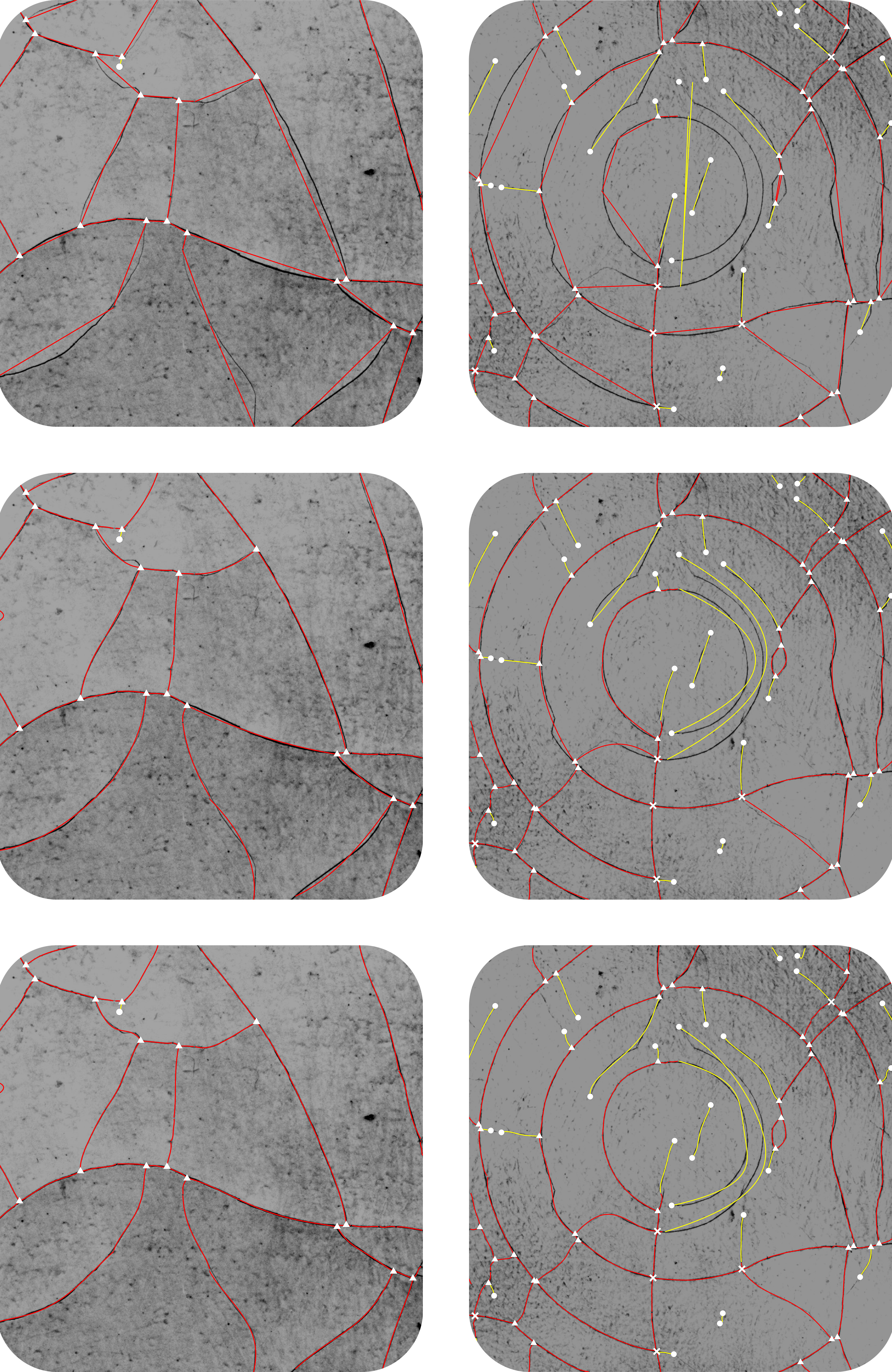}
  \end{center}
  \caption{Fracture shape reconstruction. Top row -- $n=1$, middle row -- $n=2$, bottom row –- $n=4$. An input gray-scale image is on the background. Illustrated on XVIII century French canvas paintings (left -- author NA, right -- Francois Boucher, "Venus asks Vulcan for the arms of Aeneas", 1757).}
  \label{fig:fitting}
\end{figure}

Conventional approaches of craquelure characterization omit most of the information of fractures shape and extract only their length and/or histogram of orientation, which is a serious shortcoming for applications like documentation and reconstruction. Very few algorithms \cite{abas2004}\cite{el2014development} go deeper and use chain codes as shape descriptors. However, in the case of thin lines, storing chain codes is almost identical to storing each pixel index or using the image as is. For objects with unit area, chain codes present no benefit in compression while also being sensitive to scaling and re-sampling. 

Further, we present a unified model which corrects these shortcomings and provides additional information to the basic linear approximation. The core of our model is a polynomial fitting of each crack. In this way, coordinates of each pixel within a crack are considered as a pair of values $(x_i, y_i)$ (Eq. 2) which can be exactly described by mathematical function $f^*(x_i)=y_i$. Function $f^*$ may be of any complexity, so we propose to simplify it using polynomial fitting with a strict constraint on the first and the last points (Eq. 3--4). 
\begin{equation}
    \{\mathcal{I}_ij \in (\mathcal{C}\cup\mathcal{N})\} \rightarrow \{(x_i,y_i)\in R \}
\end{equation}
\begin{equation}
\begin{tabular}{c}
    $\forall(x_i,y_i)\in \mathcal{C}:f(x_i,n) \simeq y_i \quad | $\\
    $\quad\forall(x_i,y_i)\in \mathcal{N}, \forall n\in R :f(x_i,n) = y_i$
\end{tabular}
\end{equation}
\begin{equation}
    f(x,n)=\sum_{k=0}^n \alpha_k x^k
\end{equation}

The order of a polynomial $n$ plays role of a parameter which defines the precision of reconstruction and number of stored values $\alpha_k$ (equals $n+1$). Figure \ref{fig:fitting} illustrates how even the "cheapest" fitting by a polynomial of the second order improves the reconstruction of cracks while storing only three values per edge. Experimentally, we observed that $n\leq4$ is sufficient to capture the shapes of the majority of the cracks. This allows to reduce storage space by 90-99\% in comparison to storing pixel locations or chain codes. 

Polynomial model is also efficient in a number of other applications. For example, edge orientation can be extracted as a tangent to a curve $c$, that is a first derivative of $f(x,n)$ which can easily be found analytically as $f'(x,n)=\sum_{k=1}^n k\alpha_kx^{k-1}$, or just be equal to $\alpha_1$ in linear approximation. Similarly, the curvature of a crack can be described by the absolute value of second derivative $f''(x,n)=\sum_{k=2}^n k(k-1)\alpha_k x^{k-2}$ , or just $|2\alpha_2|$ in the parabolic model.

\subsection{Properties of a network as a whole}
The features of edges and nodes described in the previous paragraphs allow to compute statistic on a whole network which then can be used as a feature vector to differentiate a given graph from a set of graphs $\{G_1,\ldots,G_N \}\in\mathcal{G}$. Such a vector may contain various hand-crafted features. Table \ref{tab:stat-features} lists the ones used in our experiments. Although this list is not exhaustive and may be easily extended with custom features or ones from previous works using the information extracted by the proposed tool. It is worth noting that we do not count closed "islands" because it is redundant when density of the nodes is accompanied by their topography. 

\begin{table}[b]
\begin{center}
\begin{tabular}{|c|p{6.2cm}|}
\hline
Notation & Feature description \\ 
\hline\hline
$\theta_\sigma$ & orientation uniformity (standard deviation of the orientation histogram) \\
$L_\mu$ & mean cracks length \\
$L_\sigma$ & standard deviation of cracks length \\
$\mathcal{N}_{/S}$ & nodes density \\
$\mathcal{N}_{O}$ & fraction of O-type nodes \\
$\mathcal{N}_{Y}$ & fraction of Y-type nodes \\
$\mathcal{N}_{X}$ & fraction of X-type nodes \\
$\mathcal{C}_{/\mathcal{N}}$ & edges to junctions ratio \\
$d_\mu^2$ & curvature as a mean of the absolute value of the second derivative \\
\hline
\end{tabular}
\end{center}
\caption{The selected statistical features.}
\label{tab:stat-features}
\end{table}

\begin{figure*}[t]
\begin{center}
  \includegraphics[width=\linewidth]{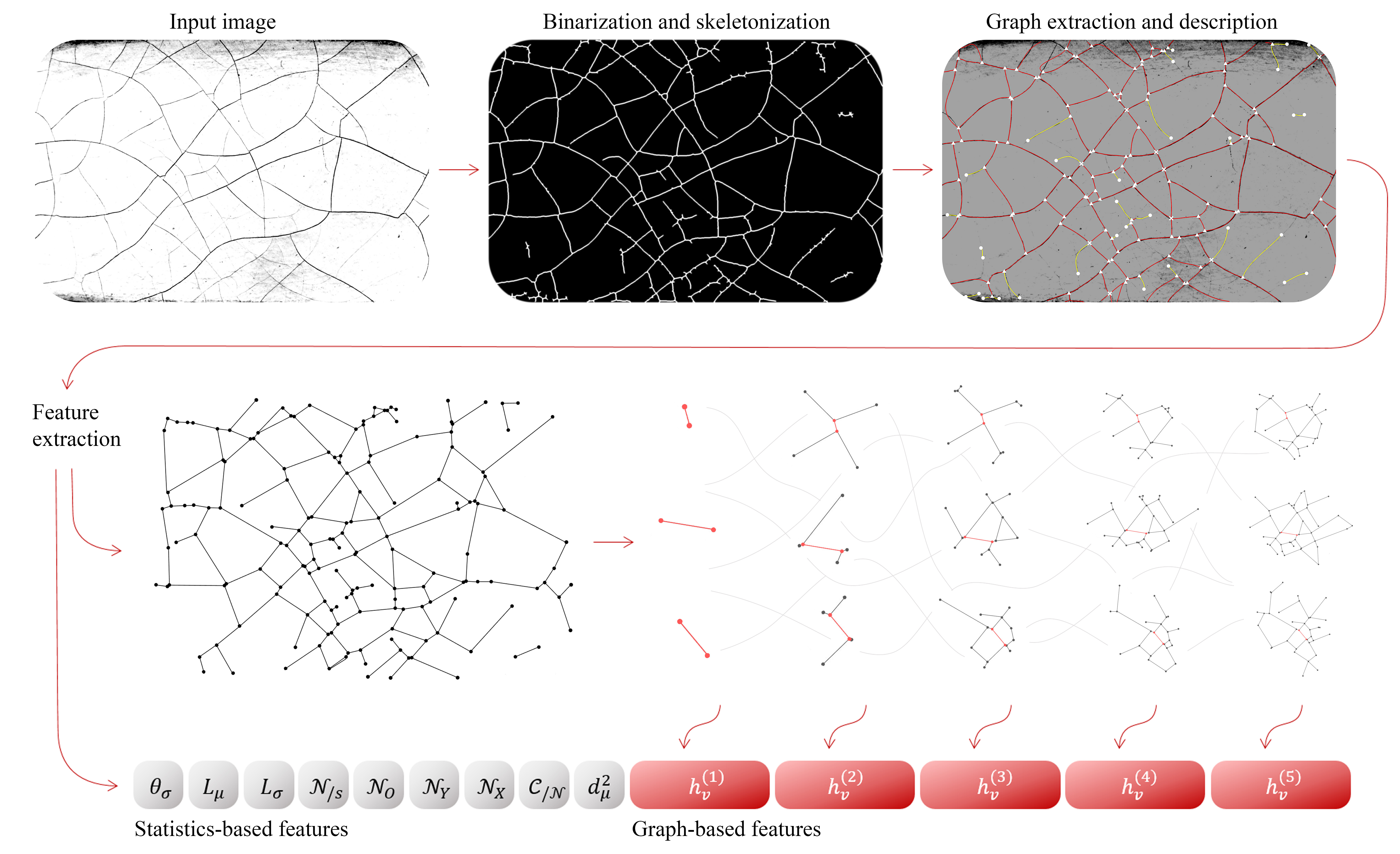}
  \end{center}
  \caption{Schematic representation of the proposed framework. The sample used for the illustration: Jean-Sim\'eon Chardin, "The House of Cards", 1737, oil on canvas.}
  \label{fig:crack-framework}
\end{figure*}

The described approach gives a statistical description of each element (node or crack) independently of its neighborhood or connections (similarly to previous works) and does not use the structure of the graph itself. Nevertheless, we hypothesize that geometrical structure of the extracted graph (regardless of the nodes' location and links' length and shape) contains useful information which is sufficient to characterize the pattern. In order to prove this, we feed unweighted non-directional graph (just indices of nodes and links between them) to a Graph Neural Network and use it as a feature extractor. Such operation is fully independent of statistical processing and when combined, they allow to obtain the most complete and comprehensive characterization of a craquelure (Fig. \ref{fig:crack-framework}).

\subsubsection{Graph Neural Networks}
Graph Neural Networks (GNNs) became the state of the art in the task of graph and node classification due to their revolutionary representational ability. GNNs follow a neighborhood aggregation scheme, where the representation vector of a node is computed by recursively aggregating and transforming representation vectors of its neighboring nodes \cite{xu2018how}. $k$ iterations of such process results in capturing structural information from all the nodes within $k$-hop neighborhood. The obtained node-wise feature representation can be used for classification of nodes or can be pooled in a global representation of a graph. 

There are a number of powerful GNN architectures, for example, Graph Convolutional NN \cite{kipf2016semi}, GraphSAGE \cite{hamilton2017inductive}, Graph Attention Model \cite{lee2018graph}, and others, but all of them share the same aggregate-combine framework (Eq. 5) and differ in the choice of $AGGREGATE^{(k)}$ and $COMBINE^{(k)}$ functions:

\begin{equation}
\begin{tabular}{c}
     $h_v^{(k)}=COMBINE^{(k)} (h_v^{(k-1)},\, a_v^{(k)}) \quad,$ \\
     $a_v^{(k)} = AGGREGATE^{(k)}\Big(\Big\{ h_{v^*}^{(k-1)},\, v^*\in V^*(v)\Big\}\Big) $
\end{tabular}
\end{equation}
where $h_v^{(k)}$ is a hidden vector-representation of a node $v\in V$ at the $k$-th layer, and $a_v^{(k)}$ is a result of the aggregation of feature-vectors from a neighborhood $V^*(v)$.

\begin{figure*}[b!]
\begin{center}
  \includegraphics[width=\linewidth]{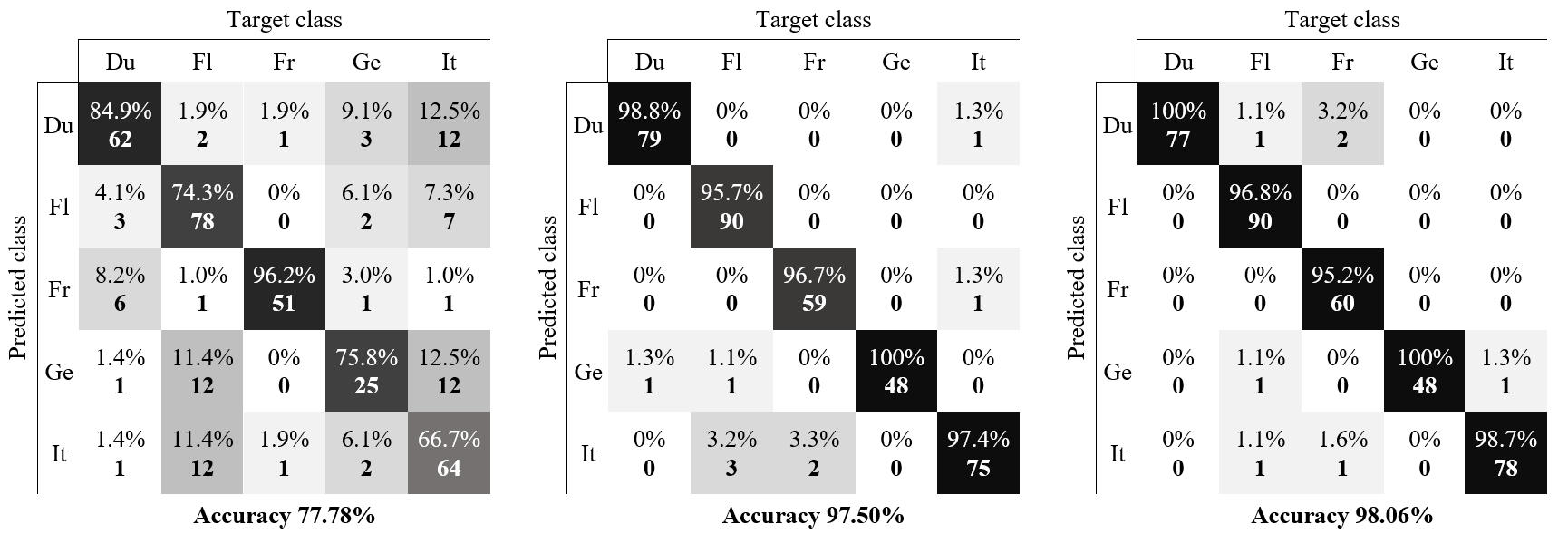}
  \end{center}
  \caption{Confusion matrices of the proposed classifiers. Left to right: statistical features, graph features, both features types.}
  \label{fig:accuracy}
\end{figure*}

Xu \etal \cite{xu2018how} (Lemma 2) define upper-bound of distinguishing ability of GNNs (mapping different graphs to different embeddings) to be as powerful as Weisfeiler-Lehman graph isomorphism test \cite{1968}. In our experiments, we utilize maximally powerful architecture (proof: Xu \etal \cite{xu2018how}, Corollary 6) called Graph Isomorphism Network which reaches this upper-bound. Therefore, a node representation is updated as:
\begin{equation}
  h_v^{(k)}=MLP^{(k)}\Bigg((1-\varepsilon^{(k)})\cdot h_v^{(k-1)}+\sum_{v^*\in V^*(v)} h_{v^*}^{(k-1)}\Bigg)  
\end{equation}
Thus, we use a 5-layer GIN, with a 2-layer Multi-Layer Perceptron ($MLP$) in each layer, to extract feature vector of a graph. The model has to be trained using labeled data in order to extract useful features. The last layer generates a feature vector $h^{(5)}$ of length 64 which theoretically covers outputs of all the previous layers. However, we find beneficial to extract hidden representation from all the layers $\{h^{(1)},\ldots,h^{(5)} \}$ and concatenate them in one longer vector. This allows to cover different sizes of a "receptive field" around a node. Consequently, we obtain a numerical description of the input graph without using any hand-crafted features. Output vector can also be combined with a statistical description generated at the previous step via simple concatenation (Fig. \ref{fig:crack-framework}).

\section{Experimental}

\subsection{Dataset}
The dataset used consists of 36 high-resolution gray-scale photographs of Italian, French, Flemish, German, and Dutch panels and canvas paintings of XIV--XVIII centuries \cite{10.2307/1506653} and was received directly from Dr. Spike Bucklow (Hamilton Kerr Institute, University of Cambridge) who originally collected the dataset. Each image has a resolution of $1181\times1772$ pixels and captures spatial region of $3\times5$ cm to $12\times20$ cm. A high density of the captured craquelure allowed to augment the dataset by random cropping each image into 10 smaller patches followed by random horizontal and vertical reflections. In result, we obtained 360 images of crack networks labeled by their geographical origin. 

\subsection{Setup}
The descriptive power of the proposed approach was evaluated on the task of classification of craquelure patterns in 5 classes. The existing approaches by Bucklow \cite{10.2307/1506653}, Abas \cite{abas2004}, and El-Youssef \textit{et al.} \cite{el2014development} are used for comparison of the results. The algorithms, source code of which is not available, were reconstructed according to their description. 10-fold cross-validation was used to minimize the influence of random train-test partition. 

Since the raw output of our algorithm is a feature vector, we used basic SVM-model with linear kernel as a classifier (may be interchanged with LDA, MLP, random forest, \textit{etc.}). In order to demonstrate benefit from each modality (or absence thereof), we analyzed the performance of the classifiers trained on statistical features and graph structural features separately as well as their combination. 

\begin{table}[t]
\begin{center}
\begin{small}
\begin{tabular}{|l|lcc|}
\hline 
\begin{tabular}{c}
     Author(s) \\
     of the method
\end{tabular}    & Features  & Classifier  & Accuracy \\ 
\hline\hline 
\multirow{2}{*}{Bucklow}       & \multirow{2}{*}{\begin{tabular}[c]{@{}l@{}}Bezier curves\end{tabular}} & MLP  & 75.30 \%  \\  &   & LDA   & 83.89 \%  \\  \hline\hline 
Abas  & \begin{tabular}[c]{@{}l@{}}Stat. features\end{tabular}  & k-NN  & 66.48 \%   \\ \hline\hline 
El-Youssef \etal  & \begin{tabular}[c]{@{}l@{}}Stat. features\end{tabular}  & LDA   & 70.15 \%  \\  \hline\hline 
\multirow{4}{*}{\textbf{ours}} & \textbf{Stat. features}                  & \multirow{4}{*}{\textbf{SVM}} & \textbf{77.78 \%}  \\
  & \textbf{GNN features}     &   & \textbf{97.50 \%}  \\
  & \textbf{GNN features+}   &   & \multirow{2}{*}{\textbf{98.06 \%}} \\  & \textbf{Stat. features}  &   &  \\
\hline
\end{tabular} 
\end{small}
\end{center}
\caption{The results of craquelure classification.}
\label{tab:classif}
\end{table}

\subsection{Results}

Results of classification are presented in Table \ref{tab:classif}. Figure~\ref{fig:accuracy} presents confusion matrices for more detailed analysis of obtained values. 

It may be seen that the classifier trained on statistical features is on a par with previous works, whereas the use of features extracted from GNN allows to outperform all the existing methods by a large margin and to achieve almost perfect accuracy. Adding the 9 statistical features helps additionally increase the score by 0.56\% and illustrates the mutual benefit of different properties of a crack pattern for its description.

It is interesting that while separately trained models misclassify Italian craquelure as Dutch, the model trained on both features does not make such an error. Also, it is worth reminding that GNN does not "know" any properties of nodes or edges between them, thus, it is very remarkable that such high performance is achieved based only on the geometrical structure of a network. Moreover, graph-representation is absolutely robust to any kind of geometrical distortions. This was not discussed before and we believe it will be a promising topic for future research.

\section{Conclusions}
In this work, we demonstrate a novel approach to the description of craquelure patterns. Firstly, we process a binarized image of a craquelure to extract a graph-like structure. Then, we train Graph Neural Network and use it as feature-extractor from a graph. Hidden representations from layers of different depth are used altogether in order to cover different sizes of a ``receptive field" around a node. At the same time, we extract the classical characteristics of the cracks and complement them by the proposed description of node topology. In the result, we obtain a powerful model for craquelure description which demonstrates state-of-the-art accuracy on a task of craquelure classification.

{\small
\bibliographystyle{ieee}
\bibliography{refs}
}

\end{document}